\documentclass{article}

\usepackage[nonatbib,preprint]{neurips_data_2022}

\usepackage[utf8]{inputenc} 
\usepackage[T1]{fontenc}    
\usepackage{hyperref}       
\usepackage{url}            
\usepackage{booktabs}       
\usepackage{amsfonts}       
\usepackage{nicefrac}       
\usepackage{microtype}      
\usepackage{xcolor}         
\usepackage{graphicx}
\usepackage{multirow}

\newcommand{\etal}{\emph{et al. }}

\title{GTAV-NightRain: Photometric Realistic Large-scale Dataset for Night-time Rain Streak Removal}

\author{%
  Fan Zhang \\
  Beijing Institute of Technology \\  
  \And
  \qquad Shaodi You \\
  \quad University of Amsterdam \\  
  \And
  Yu Li \\
  International Digital Economy Academy \\  
  \And 
  Ying Fu \\
  Beijing Institute of Technology \\
}

\begin{document}

\maketitle
\begin{figure}[ht]\small
	\centering
	\includegraphics[width=1\linewidth]{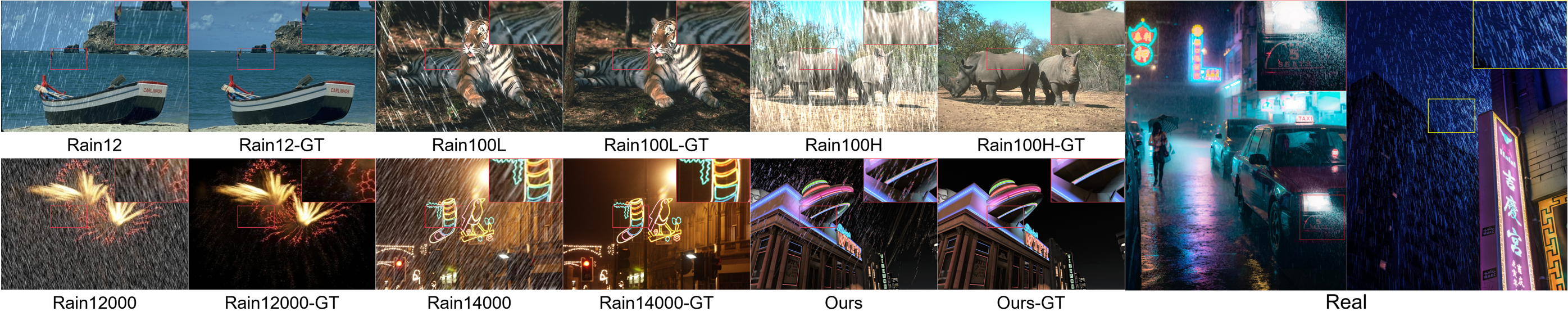}
	\caption{
	Comparisons between existing datasets and our GTAV-NightRain dataset. Rain streaks look unrealistic in the first row because the rain is superposed on 2D images of sunny days without interaction with the scene illumination and structure.
	For night scenes, they are also based on 2D superposing and their rain does not have real interaction with the scenes either.
    Rain streaks in our dataset spread in 3D space and interact with light sources both geometrically and photoelectrically, making them more realistic. Real images from the Internet are included for comparison.
	}
	\label{fig:overview}
\end{figure}

\begin{abstract}
  Rain is transparent, which reflects and refracts light in the scene to the camera. In outdoor vision, rain, especially rain streaks degrade visibility and therefore need to be removed.
  In existing rain streak removal datasets, although density, scale, direction and intensity have been considered, transparency is not fully taken into account.
  This problem is particularly serious in night scenes, where the appearance of rain largely depends on the interaction with scene illuminations and changes drastically on different positions within the image.
  This is problematic, because unrealistic dataset causes serious domain bias.
  In this paper, we propose GTAV-NightRain dataset, which is a large-scale synthetic night-time rain streak removal dataset. Unlike existing datasets,
  by using 3D computer graphic platform (namely GTA V), we are allowed to infer the three dimensional interaction between rain and illuminations, which insures the photometric realness.
  Current release of the dataset contains 12,860 HD rainy images and 1,286 corresponding HD ground truth images in diversified night scenes.
  A systematic benchmark and analysis are provided along with the dataset to inspire further research.

\end{abstract}

\section{Introduction}

Rain reflects and refracts light and causes visibility degradation, making it necessary to remove rain streaks from rainy images. Deraining has been studied for decades and deep learning methods \cite{fu2017rain14000,li2019mpid,wang2019RealRainDataset,zhang2018rain12000} gain their popularity recently like in other research areas \cite{li2021nips,ke2021nips3,liu2021nips2,liu2021nips4,chen202nips1}. 
But they mostly rely on synthetic data for training due to practical limitations, where rain streak layer is superposed onto clean images
\cite{li2016rain12,yang2017rain100,zhang2018rain12000}. 
Synthetic datasets help mitigate the problem of data shortage considering different aspects of rain such as scale, intensity, density and accumulation.

However, 
they ignore the photometric realness. 
Because rain reflects and refracts light in the scene to the camera, its appearance can be changed greatly by environmental illuminations. This is particularly problematic in night scenes, because illumination in night scenes cannot be considered as parallel and uniform. Moreover, rain appearance varies drastically depending on its position in the 3D space.
Existing datasets do not consider this issue and superpose 2D rain layer onto clean images. In a word, their rain streaks do not interact with the environment and therefore look unrealistic.

In this paper, 
instead of layer superposing, we propose to generate data by rendering rain streaks in 3D virtual scenes. 
Specifically, we use GTA V \cite{gtav} which is a popular synthetic platform for AI in urban scenes \cite{richter2016playing,richter2017playing,krahenbuhl2018free}. We investigate into the underlying asset files and mechanism related to rain rendering in GTA V. Then we make modifications using game mods to set up a suitable platform for data collection.
This allows us to edit the existence of rain in a scene while keeping everything else unchanged. In another word, we can obtain images pairs with (input) and without rain (ground truth).

Practically, in this paper, we first introduce the geometry and photometry of rain streaks. 
Second, we describe the technical details on how to edit the assets and settings of GTA V so as to collect the data.
Third, we present our new dataset GTAV-NightRain, which contains 12,860 rainy images and 1,286 clean images of a diversified night scenes.
Finally, we benchmark to analyze the existing methods using our new dataset.
The contribution of this paper is concluded as follows:

\begin{itemize}
	\item The proposed dataset is the first to consider both geometric and photometric realness for rendering airborne rain, especially in night scenes.
	
	\item The proposed dataset is large scale with diversified urban scenes and high graphic quality, which minimizes dataset bias.
	
	\item A systematical and comprehensive benchmark is provided using the proposed dataset so as to inspire further research.
\end{itemize}

\section{Related work}
In this section, we first briefly introduce existing deraining datasets and some SOTA methods which are evaluated on our dataset. Then we also mention some work related to collecting data from GTA V.

\paragraph{Deraining datasets} There are many deraining datasets proposed these years focusing on certain rain streak properties, most of which are synthesized following \cite{garg2007vision}. Li \etal \cite{li2021survey1} and Yang \etal \cite{yang2020survey2} also reviewed the methods and datasets for rain removal. Li \etal \cite{li2016rain12} provided 12 images for test (Rain12) following Garg and Shree \cite{garg2006photorealistic}. Yang \etal \cite{yang2017rain100} synthesized rain100L and rain100H with one and five types of rain streaks for light and heavy rain cases. Fu \etal \cite{fu2017rain14000} synthesized 14000 rainy images (Rain14000) with different streak orientations and magnitudes using Photoshop on 1000 clean images from UCID dataset \cite{schaefer2003ucid}, BSD dataset \cite{arbelaez2010bsd} and Google image search and. Zhang and Patel \cite{zhang2018rain12000} presented a dataset of 12000 images (Rain1200) with rain-density labels for 3 density levels. Zhang \etal \cite{zhang2019rain800} synthesized 800 images (Rain800) by following \cite{fu2017rain14000,fu2017clearing}. Wang \etal \cite{wang2019RealRainDataset} proposed a semi-automatic method to generate clean image from real rain image sequence and constructed a large-scale dataset (SPA-data) of about 29500 rain/rain-free image pairs. Li \etal \cite{li2019nyu-outdoor} proposed NYU-Rain dataset containing 16200 images and Outdoor-Rain dataset including 10500 images with rain accumulation effects rendered with depth information. Similarly, Hu \etal \cite{hu2019raincityscapes} proposed RainCityscapes based on Cityscapes dataset \cite{cordts2016cityscapes} and the rain patches are selected from \cite{li2016rain12}. Li \etal \cite{li2019mpid} proposed a large scale dataset (MPID) covering rain streaks, raindrops and rain/mist cases and also provided annotated real images for detection task after deraining.

These datasets mainly focus on the scale, shape, direction and intensities of rain streaks and some may further consider rain accumulation and different density levels. However, photometry property is not considered for rain streak removal as frequently  as adherent raindrop removal \cite{hao2019raindrop3,you2013raindrop2,you2015raindrop1,you2014raindrop4}, even though theories on this issue have been well studied \cite{garg2006photorealistic} that long ago. Thus, we attempt to take the first spite on the dataset containing colored rain steaks caused by environmental illuminations.

\begin{table}[t]\footnotesize
	\caption{Comparsions between existing datasets and GTAV-NightRain. Our dataset is not only large in amount and resolution, but also renders rain streak layer in 3D virtual scene considering photometry property rather than superposing 2D rain layer onto clean images omitting interactions with envrionment. Note that SPA-data still contains around 5 night scenes even though it is not intended for nighttime deraining.}
	\label{tab:datasets}
	\centering
	\setlength{\tabcolsep}{0.15cm}
	\begin{tabular}{lcccccc}
		\toprule
		\multirow{2}{*}{Dataset}                        & Number&\multirow{2}{*}{Resolution}&\multirow{2}{*}{Real/Synthetic}        &Layer    &\multirow{2}{*}{Day/Night}      & Light  \\
		                                                &(\#train/\#test)                     &             &                       &Type     &      & Interactions  \\ \midrule
		Rain12 \cite{li2016rain12}                      & 12                 & 481 × 321   & Synthetic             &2D     &Day     & /     \\
		Rain100L \cite{yang2017rain100}                 & 1800/200           & 481 × 321   & Synthetic             &2D     &Day     & /     \\
		Rain100H \cite{yang2017rain100}                 & 1800/200           & 481 × 321   & Synthetic             &2D     &Day     & /     \\
		Rain14000 \cite{fu2017rain14000}                & 9100/4900          & 512 × 384   & Synthetic             &2D     &Day     & /     \\
		Rain12000 \cite{zhang2018rain12000}             & 12000/1200         & 1024 × 512  & Synthetic             &2D     &Day     & /     \\
		Rain800 \cite{zhang2019rain800}                 & 700/100            & 1024 × 384  & Synthetic             &2D     &Day     & /      \\
		NYU-Rain \cite{li2019nyu-outdoor}               & 13500/2700         & 640 × 480   & Synthetic             &2D     &Day     & /     \\
		Outdoor-Rain \cite{li2019nyu-outdoor}           & 9000/1500          & 720 × 480   & Synthetic             &2D     &Day     & /     \\
		RainCityscapes \cite{hu2019raincityscapes}      & 9432/1188          & 2048 × 1024 & Synthetic             &2D     &Day     & /     \\
		SPA-data \cite{wang2019RealRainDataset}         & 28500/1000         & 512 × 512   & Real                  &3D     &Day\&Night     & Y     \\
		MPID \cite{li2019mpid}                          & 2400/250           & 512 × 384   & Synthetic\&Real        &2D     &Day     & /     \\
		GTAV-NightRain (Ours)                           & 10000/2860         & 1920 × 1080 & Synthetic             &3D     &Night   & Y     \\ \bottomrule
	\end{tabular}
\end{table}

\paragraph{Deraining methods} Ren \etal \cite{ren2019prenet} proposed a baseline network named PReNet for single image deraining with the simple combination of ResNet and multi-stage recursion and achieved promising performance. Wang \etal \cite{wang2019RealRainDataset} proposed a novel SPatial Attentive Network (SPANet) to remove rain streaks in a local-to-global manner. Wang \etal \cite{wang2020rcdnet} designed a Rain Convolutional Dictionary Network (RCDNet) for better interpretability and integration with physical structure of general rain streaks. Deng \etal \cite{deng2020drdnet} proposed Detail-Recovery image Deraining Network (DRDNet) to remove rain streaks and recover lost details. Yi \etal \cite{yi2021spdnet} utilized residue channel prior to protect struture information and guide the training of Structure Preserving Deraining Network (SPDNet). Huang \etal \cite{huang2021moss} proposed a novel Memory-Oriented Semi-Supervised method (MOSS) for single image deraining to learn rain degradations from both labeled synthetic and unlabeled real-world data. Zamir \etal \cite{Zamir2021mprnet} presented MPRNet to progressively learn restoration functions for the degraded inputs. There were also powerful image restoration backbones in recent years which can achieve good performance on deraining, such as Restormer \cite{Zamir2021restormer}, HINet \cite{chen2021hinet} and MAXIM \cite{tu2022maxim}.

\paragraph{Dataset using GTA V} 
GTA V is a popular platform for synthetic datasets. However, few work use it for low-level vision.
Richter \cite{richter2016playing} first presented an approach to rapidly producing pixel-accurate semantic label maps for images synthesized by GTA V and further presented a new benchmark suite \cite{richter2017playing} for visual perception. 
Krahenbuhl \cite{krahenbuhl2018free} presented a framework to extract ground truth supervision from video games in real-time. Doan \etal \cite{doan2018g2d} also presented G2D that could be utilized to collect the dataset in GTA V.

\section{Prerequisite: geometry and photometry of rain}
\label{sec:theories}

Unlike most existing dataset, we start from the fundamental modeling so as to ensure realness.
In this section, we briefly introduce the necessary modeling of geometry and photometry on rain. 
Then we discuss why rendered rain in GTA V is a realistic approximation.
Finally, we describe how to modify GTA V into a platform for the proposed dataset.

\subsection{The shape of raindrop}
The shape of raindrop was well studies dozens of years ago and is commonly assumed to be spherical \cite{garg2007vision}. Raindrop tends to be spherical with small size and becomes equilibrium when the size increases due to the pressure of air. It can be modeled with a 10th order cosine distortion of a sphere \cite{beard1987rainshape, garg2007vision}:
\begin{equation}
	r(\theta)=a\left(1+\sum_{n=1}^{10} c_{n} \cos (n \theta)\right),
\end{equation}
where $a$ is the spherical radius, $c_n$ is the coefficients depending on $a$ and $\theta$ is the polar angle of elevation. According to the above equation, the raindrop with radius smaller than 1mm is nearly spherical and that with larger radius is equilibrium. Furthermore, together with the commonly used empirical distribution for raindrop size named Marshall-Palmer distribution \cite{marshall1948size, garg2007vision}, one can make the observation that raindrops with size less than 1mm take the most part when it rains. It means that the shape of raindrop can be approximated by the sphere.

\subsection{Photometry of raindrop}
\begin{figure}[t]\small
	\centering
	\setlength{\tabcolsep}{0.5cm}
	\begin{tabular}{ccc}
		\includegraphics[width=0.6\linewidth]{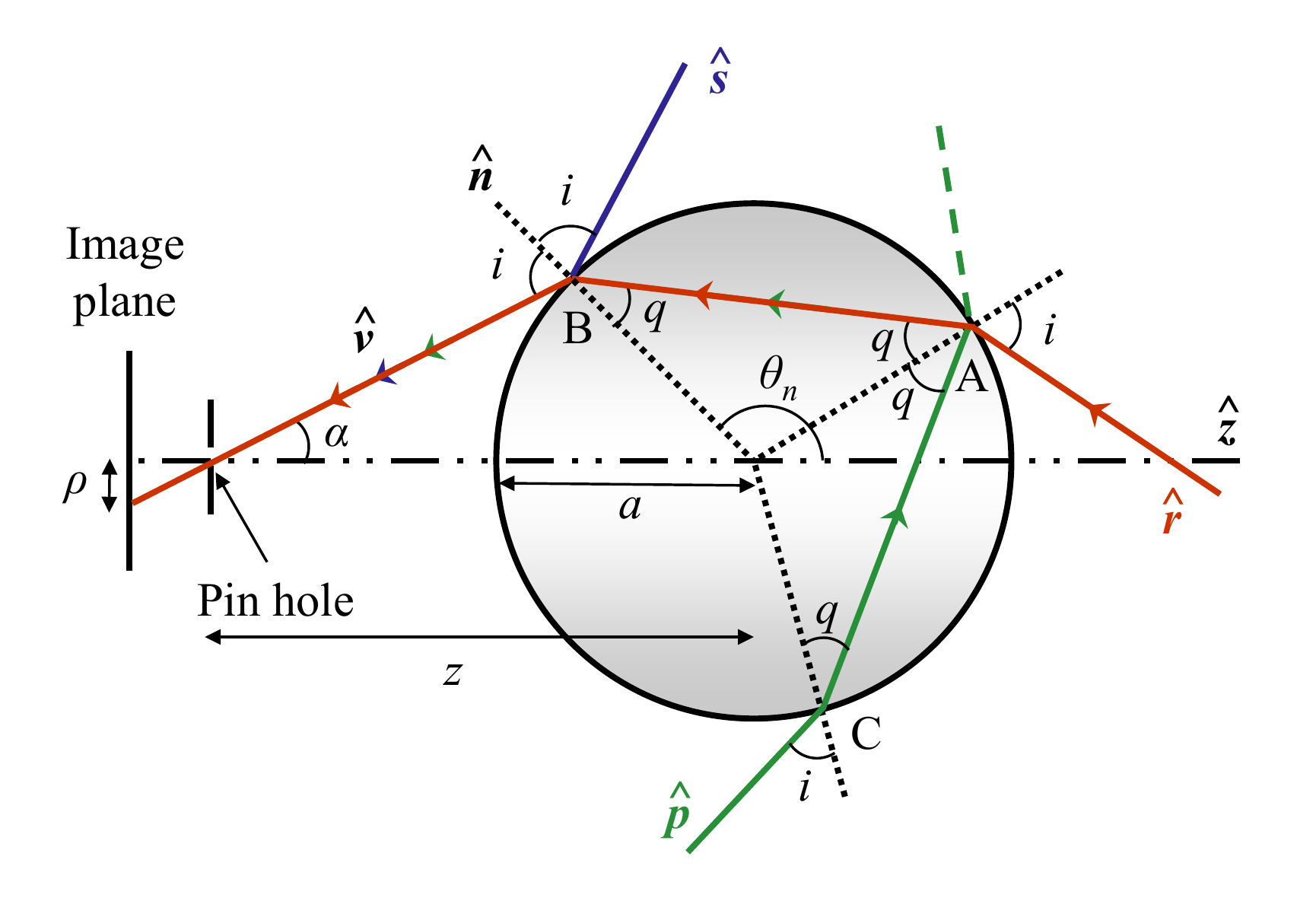} &
		\includegraphics[width=0.1\linewidth]{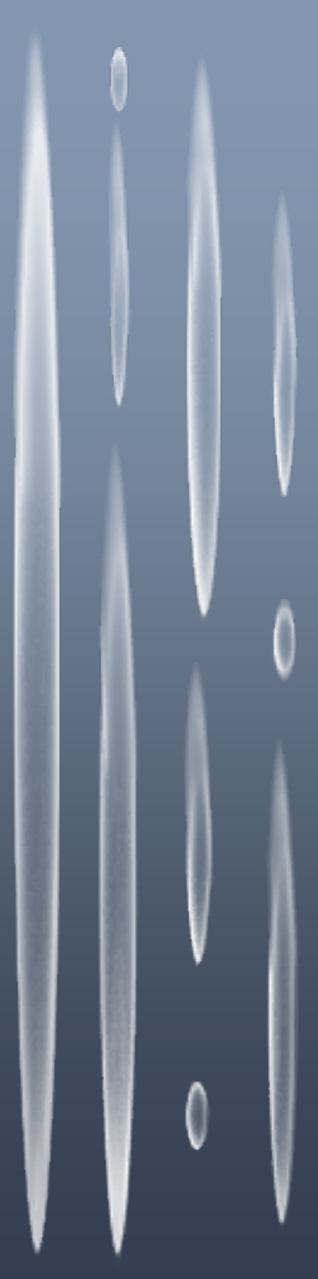} &
		\includegraphics[width=0.1\linewidth]{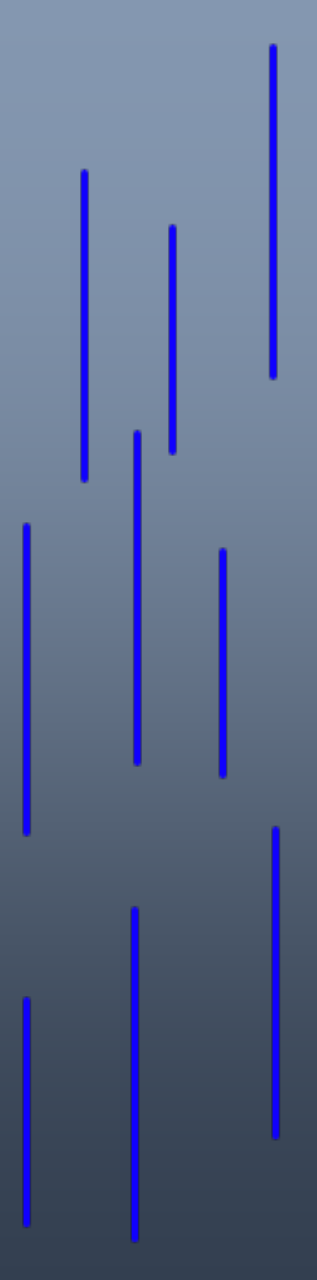} \\
		(a) & (b) & (c)
	\end{tabular}
	\caption{(a) Illustration on photometry of spherical raindrop \cite{garg2007vision}. Radiance arriving at imaging sensor consists of refracted, reflected and internally reflected light rays from environmental illumination. (b) The asset of raindrop in GTA V \cite{gtav} for \emph{set1}. (c) The asset of raindrop we draw to create \emph{set2}.}
	\label{fig:photometry}
	\vspace{-3mm}
\end{figure}

Appearance of rain purely depends on its reflection and refraction.
As shown in Fig \ref{fig:photometry} (a), considering a point B on the surface of the raindrop with a surface normal $\hat{n}$, rays from the scene $\hat{r}$, $\hat{s}$ and $\hat{p}$ are directed towards the imaging sensor via refraction, specular reflection and internal reflection, respectively. Therefore, the radiance $L(\hat{n})$ of point B can be expressed as the sum of the radiance $L_{r}$ of refracted ray, radiance $L_{s}$ of specularly reflected ray and radiance $L_{p}$ of internally reflected ray:
\begin{equation}
	L(\hat{n})=L_{r}(\hat{n})+L_{s}(\hat{n})+L_{p}(\hat{n}),
\end{equation}
due to limitation of space, 
we summarize the details in supplemental materials.
As concluded in \cite{garg2007vision}, refraction is dominating the appearance.

\subsection{Photometry of rain streaks}
While stationary raindrop can be regarded as spherical lens, falling raindrop falls fast and can not be exactly captured as it is by a camera with a normal exposure time like 30ms. Motion blur left on the image can greatly change the appearance of recorded raindrop in the scene and these trajectories are commonly called as rain streaks. The intensity of a rain streak depends on the brightness of the raindrop as well as the background scene radiance and integration time of the camera \cite{garg2007vision}.

Assuming that we observe a raining scene with a linear camera with exposure time $T$, the intensity $I_r$ at the pixel effected by a raindrop is a linear combination of the irradiance $E_b$ of the background and the irradiance $E_r$ of the raindrop:
\begin{equation}
	I_{r}=\int_{0}^{\tau} E_{r} d t+\int_{\tau}^{T} E_{b} d t.
\end{equation}
For a static background, $E_b$ can be assumed to be constant during the exposure time $T$ and thus the above equation can be expressed as:
\begin{equation}
	I_{r}=\tau \bar{E}_{r}+(T-\tau) E_{b},
\end{equation}
where $\tau$ is the time a raindrop stays at a pixel and $\bar{E}_{r}$ is the time-averaged irradiance contributed by the raindrop. For a pixel observing no raindrop, we have $I_b = E_b T$. Therefore, the change of intensity $\Delta I$ contributed by a rain streak can be expressed as:
\begin{equation}
	\Delta I=I_{r}-I_{b}=\tau\left(\bar{E}_{r}-E_{b}\right)=-\beta I_{b}+\alpha,
\end{equation}
with $\beta=\frac{\tau}{T}$, $\alpha=\tau \bar{E}_{r}$. Based on the numerical bounds and observations provided by \cite{garg2007vision}, we can know that a raindrop produces a positive change in intensity and stays at a pixel for a time far less than the integration time of a typical video camera. The change in intensities of pixels within a rain streak are linearly related to the background intensities $I_b$ and $\beta$ is in the range $0<\beta<0.039$, which means $\Delta I$ is mainly contributed by the intensity change caused by rain streak. And this is the reason why white rain streaks are directly superposed onto the image for most existing synthetic datasets.

\begin{figure}[t]\small
	\centering
	\includegraphics[width=1\linewidth]{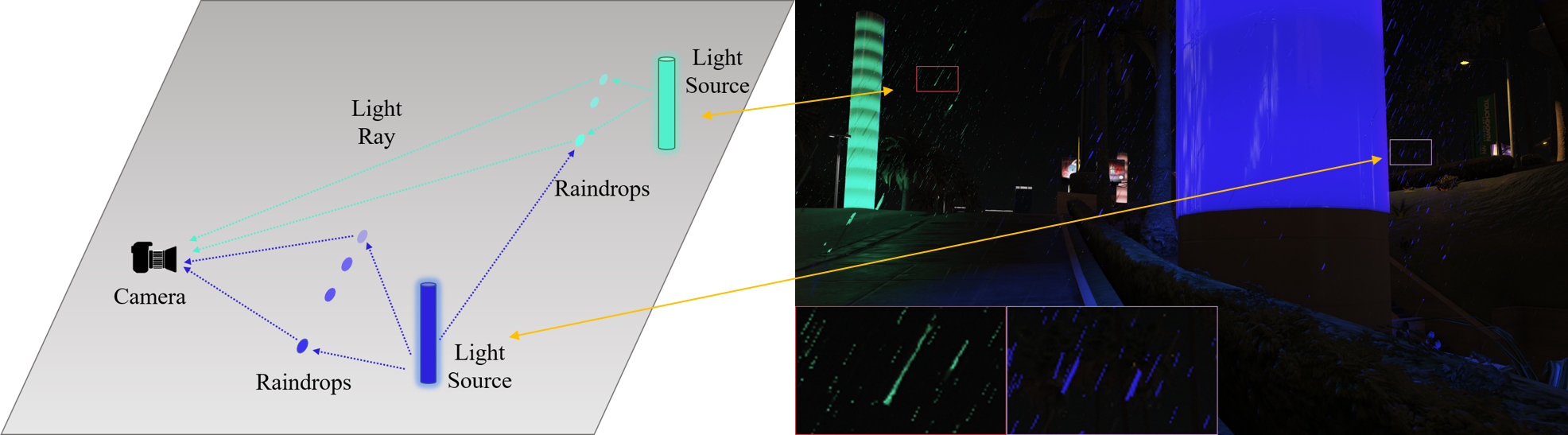}
	\caption{Illustration of photometry of rain streaks spread in 3D scene. Raindrops are responsive to dynamic and non-parallel light sources.}
	\label{fig:diagram}
	\vspace{-3mm}
\end{figure}

\section{GTAV-NightRain dataset}
\subsection{Motivation}
In real world, it is almost impossible to collect paired data with and without rain while keep everything else pixel-wisely unchanged.
Therefore, existing datasets seek solutions using synthetic rains.
As shown in Table \ref{tab:datasets}, existing datasets simply overlap an extra rain layer in 2D. However, as described in Section \ref{sec:theories} and further illustrated in Figure \ref{fig:diagram}, 
the appearance of rain streaks change drastically with space due to environmental illuminations, especially near artificial light sources in night scenes because they spread in three dimensional space and illumination in night scenes can not be considered as parallel and uniform. 

In real case, spherical raindrops are observed as rain streaks due to motion blur. While with limited rendering frame rate, artifacts occur between intervals of frames for computer game.
The rain assets are predefined to be rain-streak-like 
as shown in Fig \ref{fig:photometry} (b) and avoid this problem. 
Photometry of rain exists in GTA V and there exist light interactions between environmental illuminations and rain streaks. Raindrops in this game can also refract and reflect lights. Furthermore, these raindrops also spread across the 3D space and interact with light sources.

With all these properties, in this paper, we propose to investigate into this issue by collecting such a dataset containing dynamic and environment responsive rain streaks. Rather than superposing 2D rain streak layer onto clean images like prior work, we acquire data by rendering rain streaks in 3D virtual scenes of GTA V which has been proven to be an effective way to collect synthetic data  \cite{richter2016playing,richter2017playing,krahenbuhl2018free}. 

\subsection{Controllable conditions}
\textbf{Weather types.} In GTA V, weather changes following a predefined pseudo-random schedule. There are many weather types which are common in real life such as \emph{extra-sunny} with no clouds, \emph{clear} with a few clouds, \emph{clouds}, \emph{rain}, \emph{snow}, \emph{foggy} and \emph{etc}. Under different weather, the environment changes accordingly. Sun, clouds, fog and other environment assets also change following predefined schedule. Here we focus on the data collection for deraining, which means we need the \emph{rain} and rain-free weather like \emph{extra-sunny}, \emph{clear} or \emph{clouds}. 


\begin{figure}[t]\small
	\centering
	\includegraphics[width=1\linewidth]{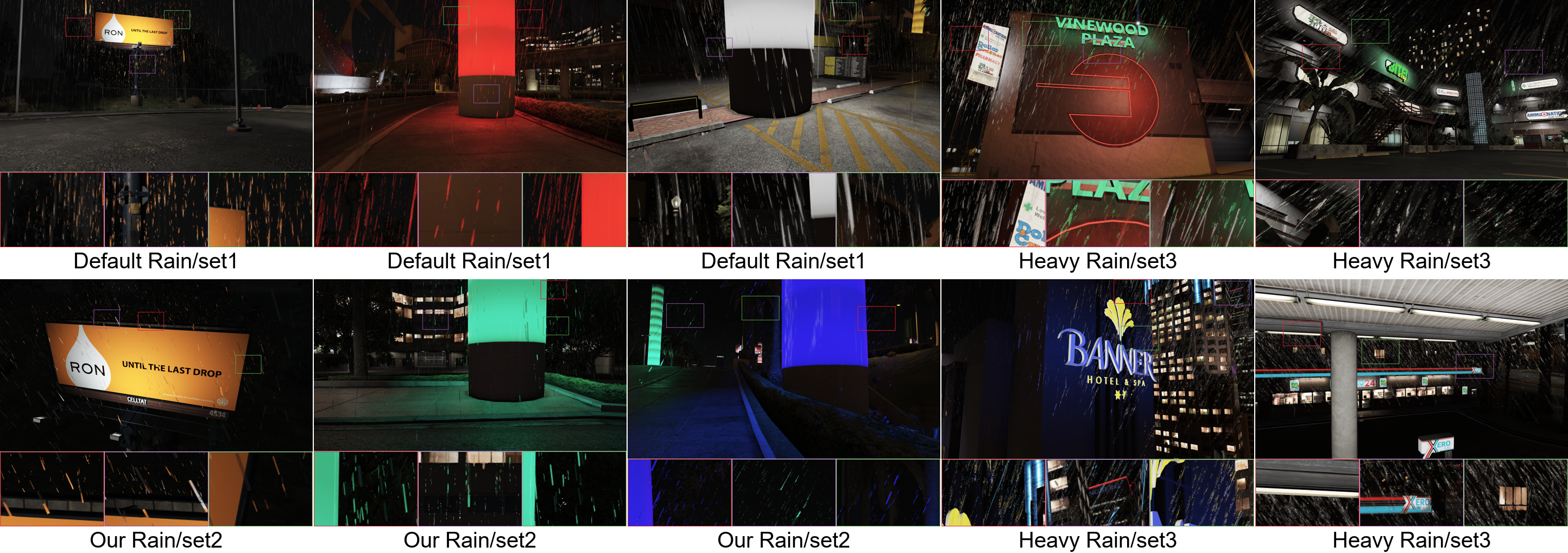}
	\caption{Example images in our GTAV-NightRain dataset. \emph{set1}, \emph{set2} and \emph{set3} correspond to splits rendered with rain shape asset by default, ours and \emph{Heavy Rain} mod, respectively. \emph{set2} intend to show the dynamic response to different lighting sources. And \emph{set3} intend to show the dynamic response to both complex lighting and structures.
	Please view in color and zoom in for details.}
	\label{fig:examples}
	\vspace{-3mm}
\end{figure}

\textbf{Environmental illuminations.} In GTA V, sun is the main light source during the day and changes greatly at different clocks such as dawn, sunset and the noon. While at night, artificial light sources account for the scene illumination, including street lights, billboards and neon lights. Moon can also contribute a lot when there is no clouds. Here we focus on night scenes where artificial light sources have great effects on the appearance of rain streaks. Thus we collect rainy and rain-free images near these local light sources.


\textbf{Rain.} In this game, raindrop shape is predefined to be rain-streak-like as shown in Fig \ref{fig:photometry} (b). Raindrops are rendered with photometry property, which means rain streaks refract or reflect lights from background. This is important but not taken into account by existing datasets. Moreover, images captured in the game contain rain streaks spread in 3D scenes, which means depth and occlusion are also considered. Rain streaks falling nearby are larger than ones far away. With occlusion, we can capture rain streaks satisfying the space arrangements in the scene by entering buildings. Moreover, the underlying rain streak shape can be substituted and we can test different raindrop shapes in the rainy scene, as shown in \ref{fig:photometry} (c). With raindrop models more fit to physical one, we can get more realistic rainy images. We can also control the density of rain for diversity.

\textbf{Puddles, ripples and splashes.} As common in real world, there will form puddles on the ground as it rains. Ripples also show up when the raindrops hit the puddle and there will be also splashes. Wetness of the environment also changes with time. We can control the existence of them or change their scales with specific parameter setting. To keep everything pixel-wisely unchanged except rain streaks, we disable the puddles, ripples and splashes before collecting data. 
A side dataset with changes on puddles, ripples and splashes will be provided in future release.

\subsection{Dataset setup}
After figuring out basic elements and mechanisms in GTA V, we are ready to collect the paired data for deraining. For our dataset, we need to collect rainy images and corresponding rain-free ground truths containing colored rain streaks in night scenes. There are several steps described as follows:
\begin{enumerate}
    \item To switch weather between \emph{rain} and \emph{clear} for example, we utilize the \emph{SimpleTrainer} mod \cite{simpletrainer}. We can also use it disable the presence of moving pedestrians, vehicles and make the character controlled invisible. It is also important to adjust the time and freeze the clock to arrange the environment settings.
    
    \item To modify the underlying files, we make use of the \emph{OpenIV} toolkit \cite{openiv} to manage mods and game files. We can change parameters of weather, lightings, assets and time schedules and also modify underlying assets like rain shape file.
    
    \item To change the appearance of rain streaks and disable puddles, we find \emph{Insane Rain} \cite{insanerain}, \emph{Real Rain} \cite{realrain} and \emph{Better Puddles} \cite{puddles} fitting our needs. We figure out internal mechanisms on how these mods work and make modifications on our own.
    
    \item The whole environment of GTA V changes following the schedule defined in \emph{timecyle} file, corresponding to different weather. To keep lightings unchanged when switching weather, we should keep parameters in the files all the same.
\end{enumerate}

For more details of modifications, please refer to the supplemental materials. To set up for data collection, we only switch weather between \emph{rain} and \emph{clear}/\emph{clouds}. The only difference is whether rain streaks exist and we keep wetness, cloud type and lightings unchanged. We disable moving pedestrians, vehicles and the character we control to avoid the change in scenes. We disable puddles, ripples and splashes to simplify the task and update the \emph{timecycle} file to keep the same lightings.

With all settings configured, rainy images and corresponding ground truth can be easily obtained with only a few operations. We collect 10 rainy images with 1 rain-free image for each scene. For a data unit of 10 rainy images, we move the character to proper location such as the front of stores and turn the camera view to proper direction like the neon light. Then we switch the weathers and capture data. In total, we test two rain shapes including the default and the one we draw. We collect 5000/500 rainy images for the training/test sets for each rain shape, making up \emph{set1} and \emph{set2}. We choose an area on purpose to collect the test set and the capture process of training data is finished away there. In addition, we also collect 1860 rainy images to form the \emph{set3} using the \emph{Insane Rain} mod with larger density level covering more scenes and select 186 images from it for the hard case. Example images are provided in Figure \ref{fig:examples}. The game is rendered in 1080P on a NVIDIA RTX 3090 GPU.

\begin{table}[t]\footnotesize
	\caption{Quantitative results of compared methods on GTAV-NightRain dataset. All the models are trained on our dataset following the default setting by authors. Higher metrics mean better performance for PSNR and SSIM. Lower metrics mean better for LPIPS.}
	\label{tab:quantitative}
	\centering
	\setlength{\tabcolsep}{0.15cm}
	\renewcommand{\arraystretch}{1.25}
	\begin{tabular}{l|c|ccc|ccc|ccc}
		\hline
		\multicolumn{1}{c|}{\multirow{2}{*}{Method}} & \multirow{2}{*}{Training} & \multicolumn{3}{c|}{Set1}                                                              & \multicolumn{3}{c|}{Set2}                                                              & \multicolumn{3}{c}{Set3}                                                               \\ \cline{3-11} 
		\multicolumn{1}{c|}{}                        &                           & \multicolumn{1}{c}{PSNR} & \multicolumn{1}{c}{SSIM} & LPIPS$\downarrow$                   & \multicolumn{1}{c}{PSNR} & \multicolumn{1}{c}{SSIM} & LPIPS$\downarrow$                    & \multicolumn{1}{c}{PSNR} & \multicolumn{1}{c}{SSIM} & LPIPS$\downarrow$                    \\ \hline
		Input                                        &                           & 33.29                    & 0.9418                   & 0.1531             & 33.68                    & 0.9550                   & 0.1360                 & 28.78                    & 0.8781                   & 0.2447                \\ \hline
		SPANet                                       & \multirow{6}{*}{Set1}     & 35.35                    & 0.9642                   & 0.0965                 & 35.11                    & 0.9661                   & 0.0993                  & 29.78                    & 0.9032                   & 0.2107                   \\
		PReNet                                       &                           & \underline{38.73}                    & \underline{0.9799}                   & \underline{0.0365}                  & 36.86                    & \underline{0.9739}                   & 0.0606                  & 32.49                    & \underline{0.9426}                   & \underline{0.1073}                \\
		DRDNet                                       &                           & 35.61                    & 0.9595                   & 0.1051               & 35.30                    & 0.9631                   & 0.0975          & 31.39                    & 0.9138                   & 0.1874                \\
		RCDNet                                       &                           & 38.69                    & 0.9781                   & 0.0404                & \underline{37.40}                    & 0.9736                   & \underline{0.0574}               & \underline{32.76}                    & 0.9369                   & 0.1147               \\
		SPDNet                                       &                           & 37.56                    & 0.9729                   & 0.0585                 & 35.96                    & 0.9687                   & 0.0769                & 32.26                    & 0.9334                   & 0.1343             \\
		MPRNet                                       &                           & \textbf{39.70}           & \textbf{0.9836}          & \textbf{0.0287}  & \textbf{37.59}           & \textbf{0.9767}          & \textbf{0.0500}        & \textbf{33.74}           & \textbf{0.9521}          & \textbf{0.0854} \\ \hline
		SPANet                                       & \multirow{6}{*}{Set2}     & 34.26                    & 0.9555                   & 0.1063                 & 34.63                    & 0.9649                   & 0.0832                   & 30.12                    & 0.9131                   & 0.1824                   \\
		PReNet                                       &                           & \underline{36.62}                    & \underline{0.9713}                   & \underline{0.0627}                  & \underline{38.45}                    & \underline{0.9789}                   & \underline{0.0323}          & 32.28                    & \underline{0.9345}                   & \underline{0.1156}          \\
		DRDNet                                       &                           & 35.14                    & 0.9564                   & 0.1128                 & 35.63                    & 0.9647                   & 0.0894                 & 31.62                    & 0.9116                   & 0.1801                \\
		RCDNet                                       &                           & 36.21                    & 0.9654                   & 0.0780                & 37.31                    & 0.9732                   & 0.0531                 & 31.94                    & 0.9233                   & 0.1399                \\
		SPDNet                                       &                           & 36.15                    & 0.9647                   & 0.0800          & 37.29                    & 0.9729                   & 0.0528                   & \underline{32.32}                    & 0.9274                   & 0.1350                \\
		MPRNet                                       &                           & \textbf{37.38}           & \textbf{0.9759}          & \textbf{0.0476}      & \textbf{39.70}           & \textbf{0.9824}          & \textbf{0.0232}      & \textbf{32.94}           & \textbf{0.9413}          & \textbf{0.0951}     \\ \hline
	\end{tabular}
	\vspace{-3mm}
\end{table}

\section{Benchmark and results}
\subsection{Experimental setups}
Based on our GTAV-NightRain dataset, we benchmark 6 existing deraining methods, including SPANet \cite{wang2019RealRainDataset}, PReNet \cite{ren2019prenet}, RCDNet \cite{wang2020rcdnet}, DRDNet \cite{deng2020drdnet}, SPDNet \cite{yi2021spdnet} and MPRNet \cite{Zamir2021mprnet}. We follow the default training settings provided by authors and train these models on the training sets of \emph{set1} and \emph{set2}. Then we test them on test sets of three subsets and real nighttime rainy images we acquired from the Internet. Model complexity is listed in Table \ref{tab:complexity} for comparison and all experiments are performed on a server with 4 NVIDIA GTX 1080Ti GPUs.

\begin{figure}[t]\small
	\centering
	\includegraphics[width=1\linewidth]{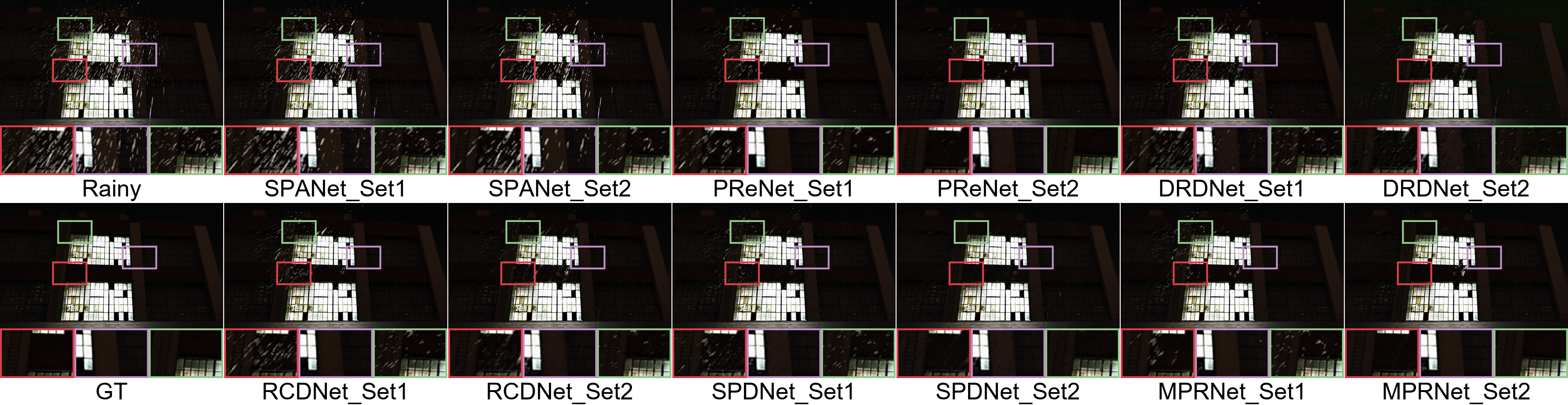}\\
	\includegraphics[width=1\linewidth]{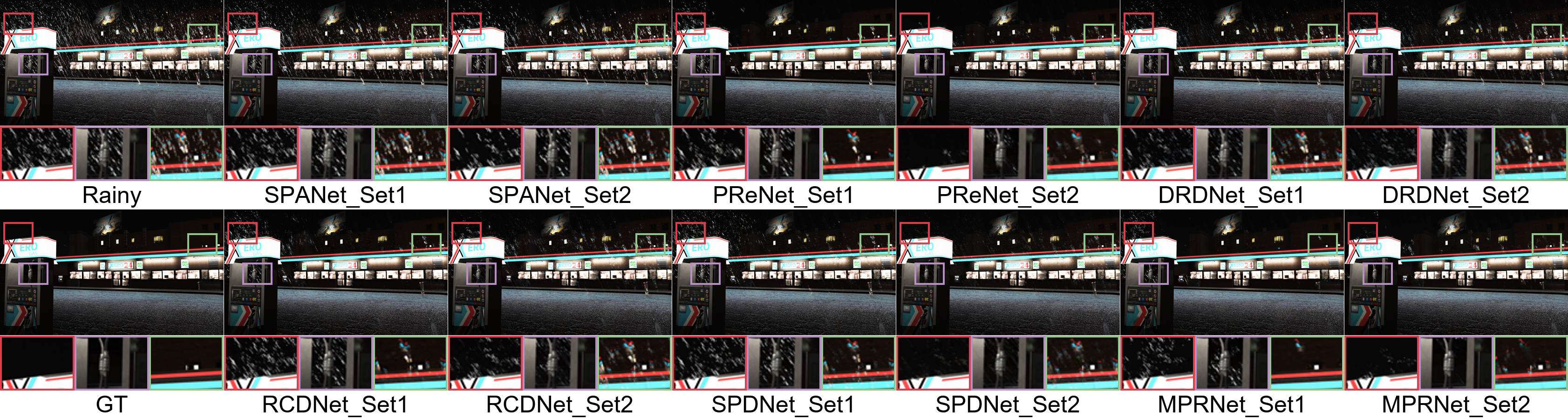}\\
	\includegraphics[width=1\linewidth]{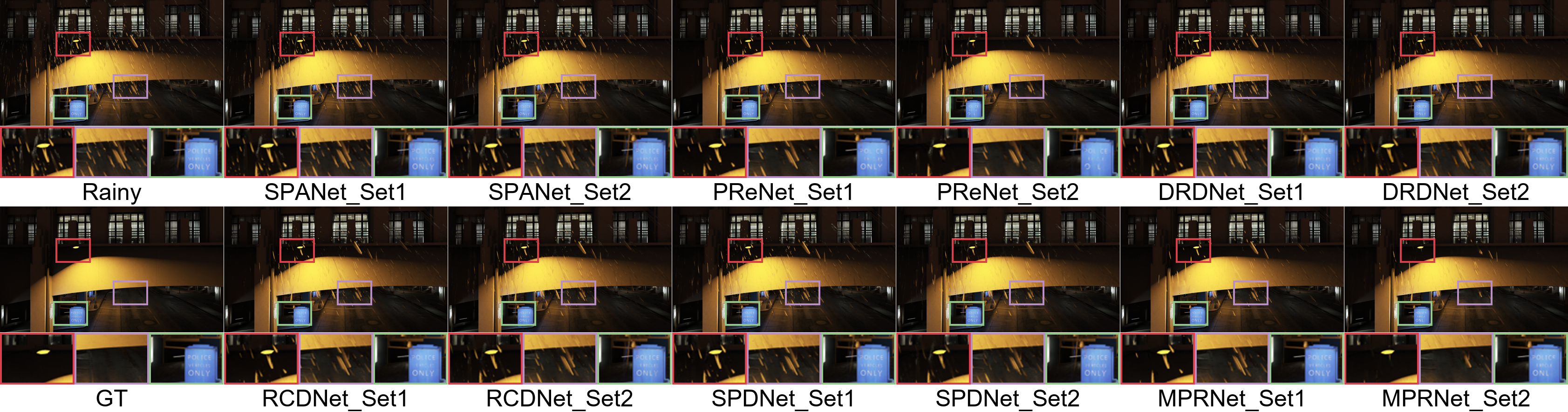}
	\caption{Qualitative results of evaluated methods on \emph{set3}. Gamma corrcetion is applied for better visibility. Please view in color and zoom in for details.}
	\label{fig:qualitative}
	\vspace{-3mm}
\end{figure}

\begin{figure}[t]\small
	\centering
	\includegraphics[width=1\linewidth]{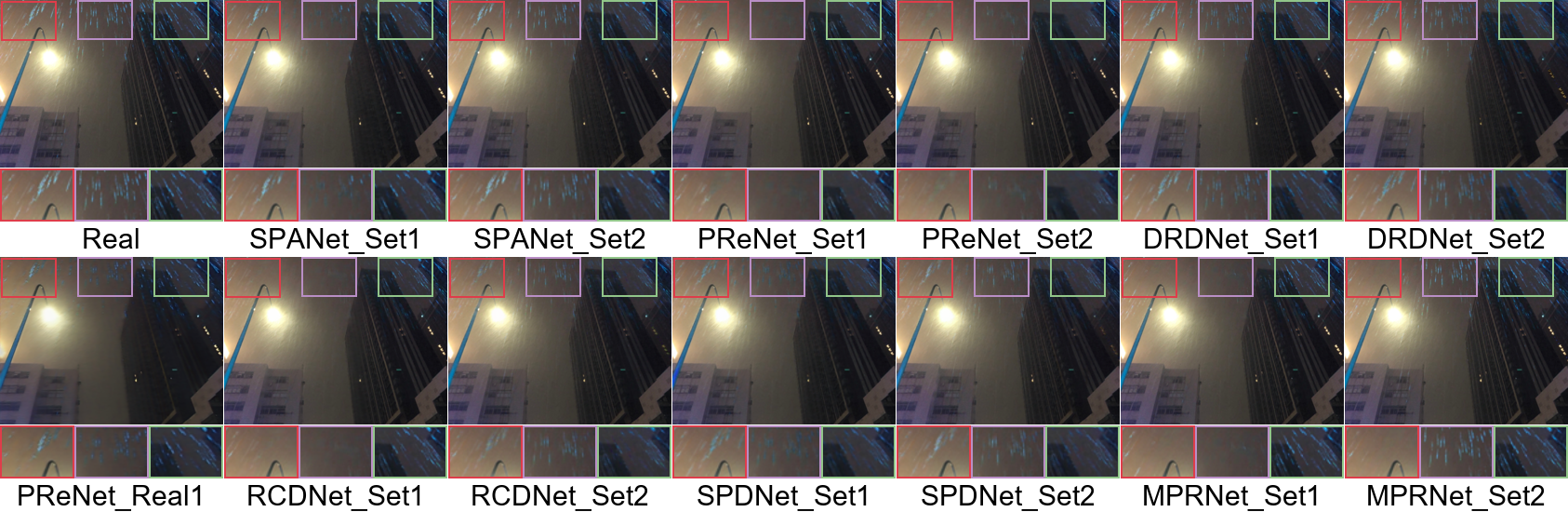}\\
	\includegraphics[width=1\linewidth]{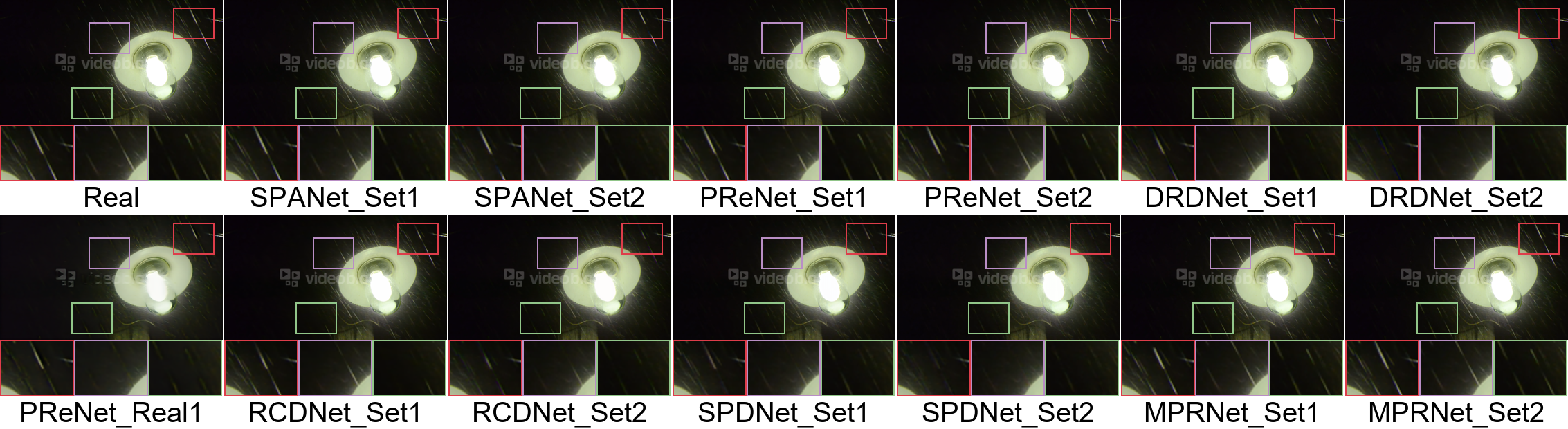}\\
	\includegraphics[width=1\linewidth]{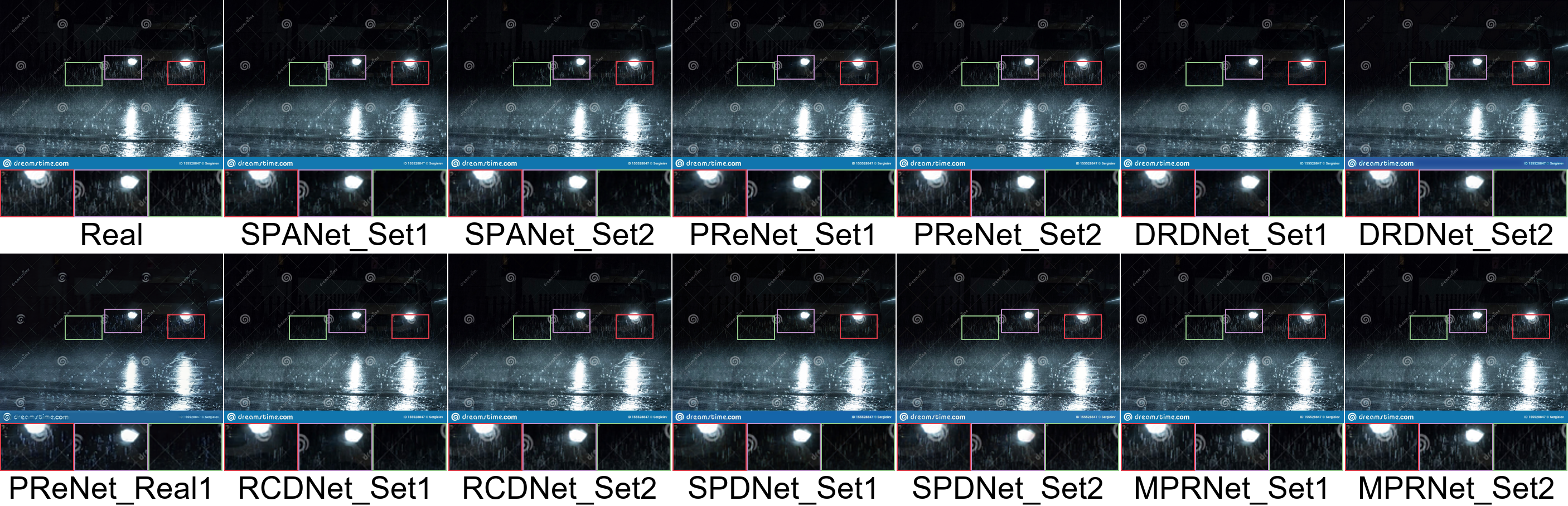}
	\caption{Qualitative results of evaluated methods on real nighttime rainy images from the Internet. The pretrained model PReNet\_Real1 is included for comparison. Please view in color and zoom in for details.}
	\label{fig:realtest}
	\vspace{-3mm}
\end{figure}

\begin{table}[t]\footnotesize
	\caption{Model complexity. The FLOPs are calculated given an test image of size 256 × 256.}
	\label{tab:complexity}
	\centering
	\setlength{\tabcolsep}{0.06cm}
	\begin{tabular}{ccccccc}
		\toprule
		\multicolumn{1}{c}{\multirow{2}{*}{Model}} & SPANet \cite{wang2019RealRainDataset}  & PReNet \cite{ren2019prenet}  & RCDNet \cite{wang2020rcdnet} & DRDNet \cite{deng2020drdnet} & SPDNet \cite{yi2021spdnet} & MPRNet \cite{Zamir2021mprnet} \\
		\multicolumn{1}{c}{}                       & (CVPR'19) & (CVPR'19) & (CVPR'20) & (CVPR'20) & (ICCV'21) & (CVPR'21) \\ \midrule
		Parameters (M)                             & 0.28    & 0.17    & 3.16    & 5.23    & 3.32    & 3.63    \\
		FLOPs (G)                                  & 36.3   & 66.4   & 195.1     & 689.8  & 96.6   & 141.5  \\ \bottomrule
	\end{tabular}
	\vspace{-5mm}
\end{table}

\subsection{Quantitative results}

We adopt commonly used PSNR and SSIM \cite{wang2004ssim} as well as perceptual metrics LPIPS \cite{zhang2018lpips} 
for performance evaluation and quantitative results are listed in Table \ref{tab:quantitative}. We can see that MPRNet \cite{wang2019RealRainDataset} trained on \emph{set1} and \emph{set2} both achieve the best performance on all test sets in terms of almost all four metrics, showing its excellent deraining ability. SPANet \cite{wang2019RealRainDataset} and DRDNet \cite{deng2020drdnet} get relatively worse metrics than other methods while PReNet \cite{ren2019prenet}, RCDNet \cite{wang2020rcdnet} and SPDNet \cite{yi2021spdnet} behave better than the former two methods. An interesting observation can be found that models with better performance (\emph{e.g.}, MPRNet \cite{Zamir2021mprnet} and PReNet \cite{ren2019prenet} drop by about 2dB in PSNR when tested on different test set) suffers more from domain gap than those with lower performance (\emph{e.g.}, SPANet \cite{wang2019RealRainDataset} and DRDNet \cite{deng2020drdnet} only drop by less than 0.5dB instead). It implies that deep models have to trade off between generalization ablity and task performance, which makes powerful models more dependent on training data. When tested on \emph{set3} which is harder, performance gap between \emph{set1} and \emph{set2} for all methods narrows to a promising extent. It means that models trained on \emph{set2} are able to handle rain streaks with appearance they've never met before. This confirms the effectiveness of the rain streak layer drawn by us and the possiblity to train more robust models by designing more diverse rain streak layers.

\subsection{Qualitative results}
Here we only provide visual results on \emph{set3} for the limited space. Please refer to supplemental material for more qualitative results. As shown in Figure \ref{fig:qualitative}, SPANet \cite{wang2019RealRainDataset} can only remove a few rain streaks while other methods achieve better results, among which MPRNet \cite{Zamir2021mprnet} behaves best. For the first test image, models trained on \emph{set2} can remove more rain than those of \emph{set1}. For the second one, models are good at removing different rain streaks, \emph{i.e.}, MPRNet\_Set1 removes colored rain streaks near the roof while MPRNet\_Set2 tends to remove more rain streaks near the fuel dispenser. It confirms the effectiveness of our rain streak layer and the possiblity to integrate real rain layers for data generation.  For the last test image, there are still some rain streaks not removed for all methods and we can find that models like PReNet\_Set2 and MPR\_Set2 tend to remove texts on the blue board.

Besides, we collect real nighttime rainy images from the Internet and existing real deraining datasets for further evaluation. We also include a pretrained model PReNet\_Real1 for comparison. As shown in Figure \ref{fig:realtest}, behaviors of all tested models vary for different real images. For the first image containing blur rain streak, PReNet\_Set1, PReNet\_Set2, RCDNet\_Set1 and MPRNet\_Set1 remove most part of rain streaks except for the right-up corner. For the second image, DRDNet\_Set2 and SPDNet\_Set2 outperform other models and remove most rain streaks. For the last rainy image, we can see that PReNet\_Set1, DRDNet\_Set1 and RCDNet\_Set1 remove rain streaks in front of headlights and DRDNet\_Set1 achieves the best result. While the pretrained model PReNet\_Real1 cannot properly handle these images. These results all prove that nighttime deraining is a challenging and both models and training data play imoprtant role in final performance on real-world rainy images. And models trained on our dataset can indeed handle these nighttime rainy images, confirming the effectiveness of our dataset.


\section{Limitations and future work}

In this paper, we propose the first nighttime deraining dataset which considers photometry of rain streaks using GTA V. Although rainy images are more similar to real world ones than existing deraining datasets, there exist limitations in our current dataset. (I) The default graphic quality of GTA V falls behind recent games which support ray tracing techniques and there is large room for improvements. (II) Due to the rendering mechanism, the game cannot be static at the very moment and we have to disable moving pedestrians and vehicles, limiting the diversity of captured scenes. (III) There inevitably exists domain gap between our synthetic dataset and real-world data thus our dataset on its own is not enough for solving nighttime deraining. However, the development of this dataset still continues and there are a lot to do in future work.

1. It is promising to update the dataset with 4K resolution and good texture mods to further enhance the realness. It is portable to newly developed game with better graphic quality. Better realness in terms of graphic quality and rendering mechanism may both narrow the gap between graphically rendered images and real captured ones.

2. We focus on deraining in night scenes in this work but illuminations also vary in day scenes. It is promising to construct full-time datasets covering both day and night scenes, enabling possible usage in hybrid tasks like low light image enhancement and deraining. Meanwhile, hybrid training of existing datasets and GTAV-NightRain may also do help to achieving better robustness of learning-based models.

3. Many assets like puddles are disabled in this version of dataset but it is promising to construct datasets with more complex rain scenes to train robust models handling more than rain streaks. 

4. Utilizing tools by Richter \etal \cite{richter2016playing} to extract assets from GPU may handle the problem of moving objects and enable more efficient way for data generation.

\section{Conclusion}
In this paper, we present a deraining dataset called GTAV-NightRain considering the photometry property of rain streak and there are 14146 images in total. Different from existing synthetic datasets, we collect data by rendering rain streaks in 3D virtual scenes in GTA V rather than superposing 2D rain streak layer on clean images. With proper modifications, we turn the game into a good platform to collect paired rainy and rain-free images, where we can easily change the weather and keep other conditions pixel-wisely unchanged. We also evaluate several SOTA deraining methods on our dataset to find what problems they will meet when facing colored rain streaks and give insights on future research interest. Portable to other newly developed games, collecting deraining dataset in such way can be benefited from the advancement of both more realistic rain modeling and better graphic quality of games, which makes the collected data better. We will also extend our dataset to day time scenes to further enrich its diversity and enhance the realess with higher resolution and graphic quality.


{\small
	\bibliographystyle{ieee_fullname}
	\bibliography{egbib}
}

\end{document}